\documentclass[letterpaper, 10 pt, journal, twoside]{IEEEtran}

\IEEEoverridecommandlockouts                              

\usepackage{amsmath} 
\usepackage{amssymb}  

\usepackage{graphicx}
\usepackage{url}
\usepackage{multirow}

\newcommand{\figref}[1]{{Fig.~\ref{#1}}}

\newcommand{\bm}[1]{\mbox{\boldmath{$#1$}}}

\title{Humanoid Loco-Manipulations \\ Pattern Generation and Stabilization Control}

\author{Masaki Murooka, Kevin Chappellet, Arnaud Tanguy, Mehdi Benallegue, \\ Iori Kumagai, Mitsuharu Morisawa, Fumio Kanehiro and Abderrahmane Kheddar
  \thanks{Manuscript received: February, 23, 2021; Accepted April, 10, 2021.}
  \thanks{This paper was recommended for publication by Editor T. Asfour upon evaluation of the Associate Editor and Reviewers' comments.} 
  \thanks{The authors are with
    CNRS-AIST JRL (Joint Robotics Laboratory), IRL and
    National Institute of Advanced Industrial Science and Technology (AIST),
    1-1-1 Umezono, Tsukuba, Ibaraki 305-8560, Japan.
    {\tt\footnotesize \{m-murooka, chappellet.kevin, arnaud.tanguy, mehdi.benallegue, iori-kumagai, m.morisawa, f-kanehiro\}@aist.go.jp}}%
  \thanks{Digital Object Identifier (DOI): see top of this page.}
}

\begin{document}

\maketitle

\markboth{IEEE Robotics and Automation Letters. Preprint Version. Accepted April, 2021}
{Murooka \MakeLowercase{\textit{et al.}}: Humanoid Loco-Manipulations Pattern Generation and Stabilization Control}

\setlength{\floatsep}{10pt}
\setlength{\textfloatsep}{8pt}
\setlength{\abovecaptionskip}{4pt}
\setlength{\abovedisplayskip}{5pt}
\setlength{\belowdisplayskip}{5pt}

\begin{abstract}
  In order for a humanoid robot to perform loco-manipulation such as moving an object while walking, it is necessary to account for sustained or alternating external forces other than ground-feet reaction, resulting from humanoid-object contact interactions. 
  In this letter, we propose a bipedal control strategy for humanoid loco-manipulation that can cope with such external forces.
  First, the basic formulas of the bipedal dynamics, i.e., linear inverted pendulum mode and divergent component of motion, are derived, taking into account the effects of external manipulation forces.
  Then, we propose a pattern generator to plan center of mass trajectories consistent with the reference trajectory of the manipulation forces, and a stabilizer to compensate for the error between desired and actual manipulation forces.
  The effectiveness of our controller is assessed both in simulation and loco-manipulation experiments with real humanoid robots.
\end{abstract}


\section{Introduction}

\IEEEPARstart{M}{oving} large and heavy objects is a hard task for humans, and is expected to be left to humanoid robots.
In such loco-manipulation tasks, humanoid robots need to walk and maintain balance while applying the force required to move the object using its `arms'.
Furthermore, the ability to achieve such a task despite the unavoidable discrepancies between planned and actual humanoid-object interaction forces, is essential to ensure robustness and reliability.

In this letter, we propose a control strategy, which can cope with external manipulation forces, for humanoid loco-manipulation.
Specifically, we extend two control layers: the pattern generator based on the preview control~\cite{PreviewControl:Kajita:ICRA2003} and the stabilizer based on the divergent component of motion (DCM) feedback control~\cite{StabilizerIntegral:Morisawa:Humanoids2012,DcmWalk:Englsberger:TRO2015,StairClimb:Caron:ICRA2019}.
These control methods are based on the usual formulas of bipedal dynamics, i.e., linear inverted pendulum mode (LIPM)~\cite{LIPM:Kajita:IROS2001} and DCM dynamics~\cite{DCM:Takenaka:IROS2009,DcmWalk:Englsberger:TRO2015}.
By re-deriving and re-examining these key formulas taking into account external manipulation forces, the proposed control can accurately handle the following two points that were ignored in the conventional controls of humanoid loco-manipulation~\cite{PushZMP:Harada:ICRA2003,Push:Takubo:ICRA2005,Push:Nishiwaki:Humanoids2006,Push:Stilman:ICRA2008,Push:Hakamata:AR2020}:
\begin{enumerate}
\item incorporating the effect of manipulation vertical forces without approximation in the walking pattern generation, and
\item explicitly compensating for the error of manipulation forces in the stabilizer.
\end{enumerate}
We show the effectiveness of the proposed control strategy through simulations and real experiments, in which humanoid robots perform various loco-manipulation tasks.

\section{Background and Contribution}

\subsection{Bipedal Walking}

A widely used framework for controlling bipedal walking consists of a pattern generator and a stabilizer.
The pattern generator is responsible for generating the trajectory of the center of mass (CoM) to track the reference ZMP trajectory. Several methods have been proposed, including preview control-based method~\cite{PreviewControl:Kajita:ICRA2003}, MPC-based method~\cite{MPCWalk:Herdt:AR2010}, and DCM-based method~\cite{DcmWalk:Englsberger:TRO2015}.
The stabilizer's role is to steer the robot's CoM to track the desired CoM trajectory generated by the pattern generator based on robot state sensor measurements (or estimators).
Feedback control based on DCM or its equivalent values is also used in the stabilizer~\cite{BestRegulator:Sugihara:ICRA2009,StabilizerIntegral:Morisawa:Humanoids2012,DcmWalk:Englsberger:TRO2015,StairClimb:Caron:ICRA2019}.
The methods in the previous studies are mostly based on LIPM~\cite{LIPM:Kajita:IROS2001} and DCM dynamics~\cite{DCM:Takenaka:IROS2009,DcmWalk:Englsberger:TRO2015}. In this letter, the pattern generator and stabilizer are extended to account for external manipulation forces in their closed-form formulas.

\subsection{Humanoid Loco-manipulation}

In previous studies, loco-manipulation such as non-prehensile manipulation (e.g., pushing~\cite{PushZMP:Harada:ICRA2003,Push:Takubo:ICRA2005,Push:Nishiwaki:Humanoids2006,Push:Stilman:ICRA2008,Push:Hakamata:AR2020} and pivoting~\cite{Pivot:Yoshida:AR2009,ManipulationStrategyDecision:Murooka:ICRA2014}) and articulated environment operation (e.g., door opening~\cite{OpenDoor:Nozawa:IROS2012}) has been achieved by life-sized humanoid robots.
In most of these studies, the pattern generator and stabilizer for bipedal walking is used as is, with an offset due to hand forces added to the ZMP or CoM~\cite{PushZMP:Harada:ICRA2003,Push:Takubo:ICRA2005,Push:Nishiwaki:Humanoids2006,Push:Stilman:ICRA2008,Push:Hakamata:AR2020}.
This corresponds to approximating the ZMP scale effect due to hand vertical forces as negligible.
In this letter, we show that this approximation can be eliminated without increasing the computational cost.
Besides, for stable humanoid loco-manipulations, it is important to deal with the errors in manipulation forces.
However, in previous studies, although manipulation contact positions were adjusted according to the measured force by impedance control~\cite{Push:Stilman:ICRA2008,Push:Nishiwaki:Humanoids2006,OpenDoor:Nozawa:IROS2012}, the manipulation force discrepancy was not considered in the stabilizer.
To the best of our knowledge, this letter is the first to formulate the CoM and ZMP adjustments in a DCM-based bipedal stabilizer with explicit consideration of the manipulation force error. In~\cite{HumanCollab:Agravante:TRO2019} external forces are accounted for, yet without dedicating an appropriate stabilizer, which resulted in a very conservative joint walking. Another shortcoming was the interaction force prediction (that was part of the MPC).


\subsection{Multi-contact Motion}

Loco-manipulation discussed in this letter can be categorized as a special case of multi-contact motion: mainly walking on two legs while manipulating an object with the hands.
Recently, multi-contact motion generation~\cite{MultiContact:Audren:IROS2014,MultiContactStabilizer:Morisawa:IROS2019} and stabilization~\cite{MultiContact:Farnioli:ICRA2015,RobustLadder:Kanazawa:IROS2015,MultiContact:Farraj:RAL2019,MultiContactStabilizer:Morisawa:IROS2019} have been proposed, including pushing operations using multi-contact controllers~\cite{MultiContact:Hiraoka:AR2021,MultiContactPush:Polverini:RAL2020}.
These methods treat arms and legs indistinguishably in planning and control, and are based on general equations of motion that do not assume bipedal dynamics modes.
However, w.r.t. bipedal walking performance, these methods still have limitations, such as higher computational costs for motion generation and longer time for stabilization to converge.
In this letter, by assuming loco-manipulation with bipedal walking, we adopt a bottom-up approach that extends the low computational cost methods based on bipedal dynamics, rather than the top-down general multi-contact motion methods with high computational cost.

%

\subsection{Overall Configuration of our Control System} \label{sec:system}

\begin{figure}[tpb]
  \begin{center}
    \includegraphics[width=1.0\columnwidth]{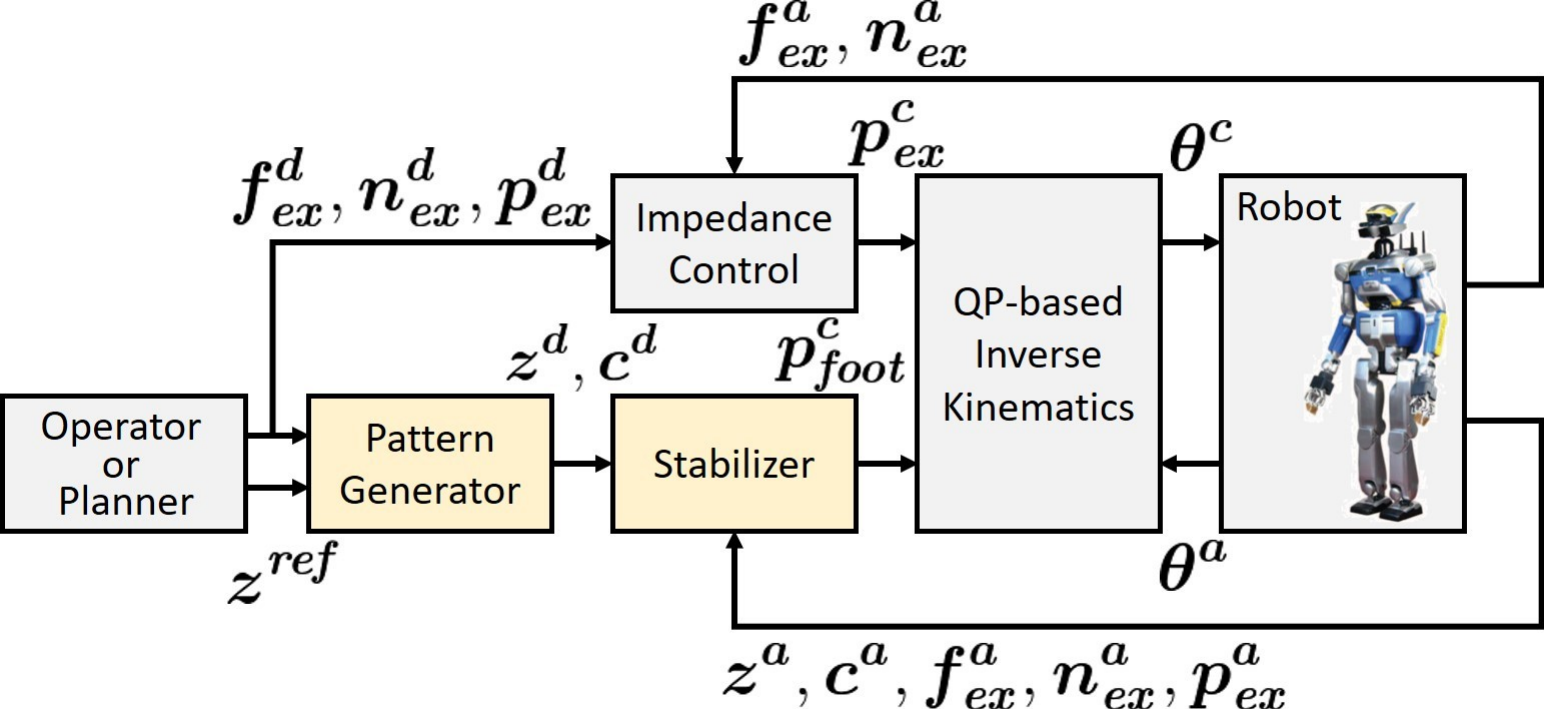}
    \caption{Overall components for humanoid loco-manipulation.
      \newline \footnotesize
      The superscripts ``d'', ``a'', and ``c'' mean the desired value, actual (measured) value, and command value, respectively.
      $\bm{\theta}$ represents the robot joint angles.
    }
    \label{fig:system}
  \end{center}
\end{figure}

\figref{fig:system} illustrates our proposed control system for humanoid loco-manipulation considering external manipulation forces (e.g., on the hands).
The main focus of this letter is on pattern generator and stabilizer.
These two main ingredients are presented in Section~\ref{sec:pg} and~\ref{sec:st}, respectively, after deriving key formulas of the bipedal dynamics with external forces in Section~\ref{sec:basic}.
Section~\ref{sec:val} describes the implementation and assessments using simulation, and Section~\ref{sec:exp} shows the application to various loco-manipulation tasks in simulation and real world.

\section{Bipedal Dynamics with External Forces} \label{sec:basic}

In this section, we derive LIPM~\cite{LIPM:Kajita:IROS2001} and DCM dynamics~\cite{DCM:Takenaka:IROS2009,DcmWalk:Englsberger:TRO2015}
that take into account external manipulation forces.
These formulas are the basis of the pattern generator and the stabilizer presented in the following sections.

\subsection{Newton-Euler Equation}

\begin{figure}[tpb]
  \begin{center}
    \includegraphics[width=0.75\columnwidth]{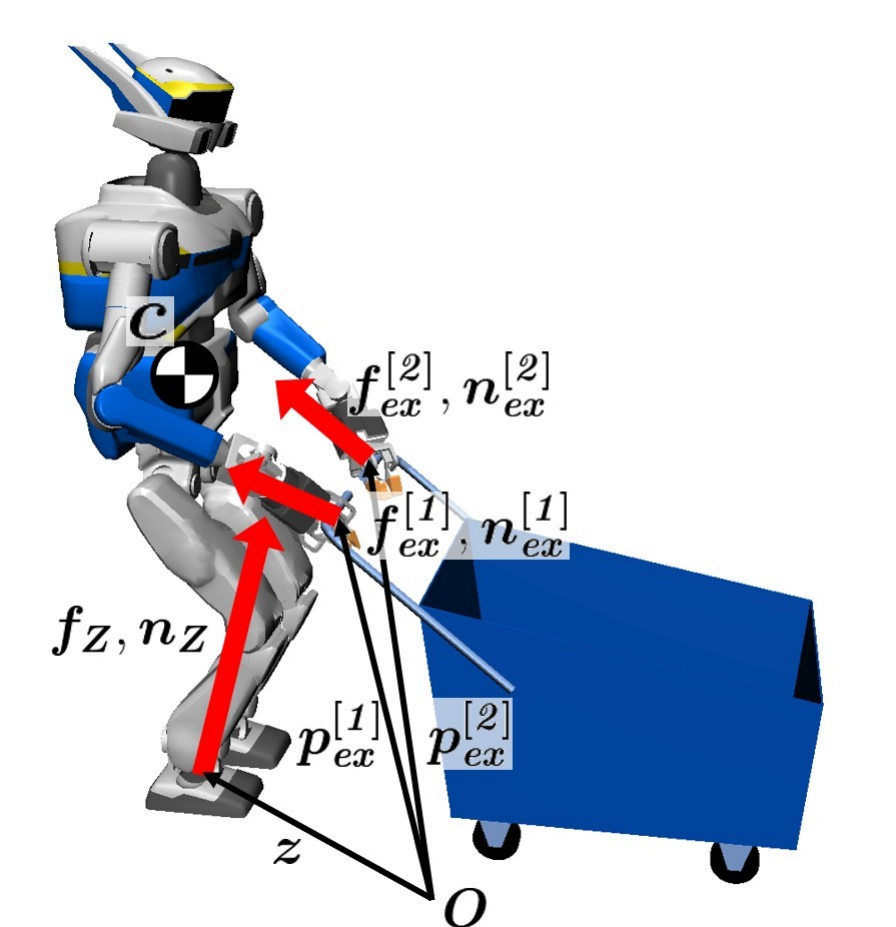}
    \caption{Centroidal dynamics in humanoid loco-manipulation case.
    }
    \label{fig:newton-euler}
  \end{center}
\end{figure}

The centroidal dynamics of a humanoid robot are governed by the Newton-Euler equation (\figref{fig:newton-euler}):
\begin{multline}
  \begin{bmatrix} m \bm{\ddot{c}} \\ \bm{\dot{L}_c} \end{bmatrix}
  =
  \begin{bmatrix} \bm{f_Z} \\ \left( \bm{z} - \bm{c} \right) \times \bm{f_Z} + \bm{n_Z} \end{bmatrix}
  -
  \begin{bmatrix} m \bm{g} \\ \bm{0} \end{bmatrix}
  \\ +
  \sum_i
  \begin{bmatrix}
    \bm{f^{[i]}_{\mathit{ex}}} \\
    \left(\bm{p^{[i]}_{\mathit{ex}}} - \bm{c}\right) \times \bm{f^{[i]}_{\mathit{ex}}} + \bm{n^{[i]}_{\mathit{ex}}}
  \end{bmatrix}
  \label{eq:newton-euler}
\end{multline}
$m \in \mathbb{R}$ is the robot mass.
$\bm{g} = [0 \ 0 \ g]^T$ is the gravitational acceleration vector.
$\bm{c} \in \mathbb{R}^3$ is the robot CoM.
$\bm{L}_c \in \mathbb{R}^3$ is the angular momentum around the CoM.
$\bm{z} \in \mathbb{R}^3$ is the ZMP.
Note that $\bm{z}$ 
is calculated directly and only from the feet forces.
$\bm{f_Z}$ and $\bm{n_Z} \in \mathbb{R}^3$ are the net force and moment around the ZMP that the robot receives through the feet.
$\bm{f^{[i]}_{\mathit{ex}}}, \bm{n^{[i]}_{\mathit{ex}}}, \bm{p^{[i]}_{\mathit{ex}}} \in \mathbb{R}^3$ are the force, moment, and position of the $i^{\text{th}}$ external contact, respectively.

One notable loco-manipulation feature that differs from the general multi-contact motion, is that the reference value of the manipulation commands ($\bm{f^{[i]}_{\mathit{ex}}}, \bm{n^{[i]}_{\mathit{ex}}}, \bm{p^{[i]}_{\mathit{ex}}}$) is given. 
For example, the hand positions can be planned from the target object trajectory, and the hands' forces and moments can be computed from the expected force required to move an object. 
Even if the expected required forces are not known \emph{a priori}, it is still possible for the robot to estimate them online from sensor measurements~\cite{Push:Stilman:ICRA2008,OpenDoor:Nozawa:IROS2012,WholebodyPush:Murooka:ICRA2015}. 
In this letter, we take full advantage of this feature and extend the conventional methods for bipedal walking, assuming application to loco-manipulation. 
Although the reference value of manipulation commands is considered given or known, we stress that the error discrepancy with the actual value is explicitly compensated in our proposed stabilizer.

\subsection{Linear Inverted Pendulum Mode}

Eliminating $\bm{f_Z}$ from \eqref{eq:newton-euler} and expressing it in component form: 

\noindent 
\begin{multline}
  \begin{bmatrix} \dot{L}_{c,x} \\ \dot{L}_{c,y} \\ \dot{L}_{c,z} \end{bmatrix}
  =
  \begin{bmatrix} z_x - c_x \\ z_y - c_y \\ z_z - c_z \end{bmatrix}
  \times
  \begin{bmatrix} m \ddot{c}_x - \sum_i f^{[i]}_{ex,x} \\ m \ddot{c}_y - \sum_i f^{[i]}_{ex,y} \\ m (\ddot{c}_z + g) - \sum_i f^{[i]}_{ex,z} \end{bmatrix}
  +
  \begin{bmatrix} n_{Z,x} \\ n_{Z,y} \\ n_{Z,z} \end{bmatrix}
  \\ +
  \sum_i \left(
  \begin{bmatrix} p^{[i]}_{ex,x} - c_x \\ p^{[i]}_{ex,y} - c_y \\ p^{[i]}_{ex,z} - c_z \end{bmatrix}
  \times
  \begin{bmatrix} f^{[i]}_{ex,x} \\ f^{[i]}_{ex,y} \\ f^{[i]}_{ex,z} \end{bmatrix}
  +
  \begin{bmatrix} n^{[i]}_{ex,x} \\ n^{[i]}_{ex,y} \\ n^{[i]}_{ex,z} \end{bmatrix}
  \right)
  \label{eq:newton-euler-element}
\end{multline}
$*_x, *_y, *_z$ are the X, Y, and Z components of $\bm{*}$, respectively.

$n_{Z,x}$ and $n_{Z,y}$ are zero from the definition of ZMP.
Assuming the horizontal components of the angular momentum rate $\dot{L}_{c,x}, \dot{L}_{c,y}$ are zero,
the X and Y components of \eqref{eq:newton-euler-element} are transformed as follows:
\begin{subequations}
\begin{align}
  & \ddot{c}_x = \omega^2 \left( c_x - \kappa z_x + \gamma_x \right) \label{eq:lipm-x} \\
  & \ddot{c}_y = \omega^2 \left( c_y - \kappa z_y + \gamma_y \right) \label{eq:lipm-y}
\end{align}
\end{subequations}
where
\addtocounter{equation}{-1}
\begin{subequations}
\setcounter{equation}{2}
\begin{align}
  & \omega = \sqrt{\frac{\ddot{c}_z + g}{c_z - z_z}} \\
  & \kappa = 1 - \frac{\sum_i f^{[i]}_{ex,z}}{\zeta}, \ \ \ \ \ \ \ \ \zeta = m \left( \ddot{c}_z + g \right) \label{eq:lipm-scale} \\
  & \gamma_x = \frac{1}{\zeta} \sum_i \left(
  \left( p^{[i]}_{ex,z} \!-\! z_z \right) f^{[i]}_{ex,x} \!-\! p^{[i]}_{ex,x} f^{[i]}_{ex,z} \!+\! n^{[i]}_{ex,y}
  \right) \label{eq:lipm-offset-x} \\
  & \gamma_y = \frac{1}{\zeta} \sum_i \left(
  \left( p^{[i]}_{ex,z} \!-\! z_z \right) f^{[i]}_{ex,y} \!-\! p^{[i]}_{ex,y} f^{[i]}_{ex,z} \!-\! n^{[i]}_{ex,x}
  \right) \label{eq:lipm-offset-y}
\end{align}
\end{subequations}

Combining \eqref{eq:lipm-x} and \eqref{eq:lipm-y}, we get the first key formula, the LIPM with external forces:
\begin{align}
  & \bm{\ddot{c}} = \omega^2 \left( \bm{c} - \kappa \bm{z} + \bm{\gamma} \right) \label{eq:lipm}
\end{align}
$\bm{c} \in \mathbb{R}^2$ is the CoM.
$\bm{z} \in \mathbb{R}^2$ is the ZMP.
The vertical component is dropped from here, and only the horizontal components are considered.
Assuming constant vertical CoM and ZMP positions, treat $\omega$ as a constant.
Compared to the conventional LIPM~\cite{LIPM:Kajita:IROS2001}, the natural frequency $\omega$ remains the same, with additional scale $\kappa$ and offset $\bm{\gamma} = [\gamma_x \ \gamma_y]^T$.
The X and Y components of the CoM and ZMP are completely separated from the other coefficients ($\omega$, $\kappa$, and $\bm{\gamma}$).
Also, the external forces only affect $\kappa$ and $\bm{\gamma}$, not $\omega$.
These features are important in dealing with external forces in a LIPM-based approach.
As a difference from the previous studies~\cite{PushZMP:Harada:ICRA2003,Push:Takubo:ICRA2005,Push:Nishiwaki:Humanoids2006,Push:Stilman:ICRA2008,Push:Hakamata:AR2020}, where \eqref{eq:newton-euler-element} is solved for ZMP; in this letter, it is organized as the second-order dynamics of the CoM, which clarifies the bipedal dynamics with external forces.

\subsection{Divergent Component of Motion}

Expressing \eqref{eq:lipm} as a state equation:
\begin{align}
  & \frac{d}{dt}
  \begin{bmatrix} \bm{c} \\ \bm{\dot{c}} \end{bmatrix}
  =
  \bm{A_c}
  \begin{bmatrix} \bm{c} \\ \bm{\dot{c}} \end{bmatrix}
  +
  \bm{B_c}
  (\kappa \bm{z} - \bm{\gamma})
  \label{eq:lipm-state-eq} \\
  & {\rm where} \ \ \ \ \bm{A_c} = \begin{bmatrix} \bm{O} & \bm{I} \\ \omega^2 \bm{I} & \bm{O} \end{bmatrix},
  \bm{B_c} = \begin{bmatrix} \bm{O} \\ - \omega^2 \bm{I} \end{bmatrix} \nonumber
\end{align}
$\bm{O}$ and $\bm{I}$ are a two-dimensional zero matrix and an identity matrix, respectively.
The system matrices $\bm{A_c}$ and $\bm{B_c}$ are the same as the conventional state equation without external forces.
The eigenvalues of $\bm{A_c}$ are $-\omega$ and $\omega$.
The eigenvectors corresponding to the eigenvalue $\omega$ are $[1 \ 0 \ \omega \ 0]^T$ and $[0 \ 1 \ 0 \ \omega]^T$, and their component is the following unstable mode called DCM~\cite{DCM:Takenaka:IROS2009,DcmWalk:Englsberger:TRO2015}:
\begin{align}
  \bm{\xi} = \bm{c} + \frac{1}{\omega} \bm{\dot{c}} \label{eq:dcm}
\end{align}
Note that even if the robot is subject to external forces, the DCM is the same as in the conventional case~\cite{DCM:Takenaka:IROS2009,DcmWalk:Englsberger:TRO2015}.
The state equation \eqref{eq:lipm-state-eq} can be expressed using DCM as follows:
\begin{align}
  \frac{d}{dt}
  \begin{bmatrix} \bm{c} \\ \bm{\xi} \end{bmatrix}
  =
  \begin{bmatrix} - \omega \bm{I} & \omega \bm{I} \\ \bm{O} & \omega \bm{I} \end{bmatrix}
  \begin{bmatrix} \bm{c} \\ \bm{\xi} \end{bmatrix}
  +
  \begin{bmatrix} \bm{O} \\ - \omega \bm{I} \end{bmatrix}
  (\kappa \bm{z} - \bm{\gamma})
  \label{eq:lipm-state-eq-with-dcm}
\end{align}

The upper part of \eqref{eq:lipm-state-eq-with-dcm} is similar to the conventional case.
Extracting the lower part of \eqref{eq:lipm-state-eq-with-dcm}, we get the second key formula, the first-order dynamics of DCM:
\begin{align}
  \bm{\dot{\xi}} = \omega (\bm{\xi} - \kappa \bm{z} + \bm{\gamma}) \label{eq:dcm-dynamics}
\end{align}
If the external forces are zero, $\kappa = 1$ and $\bm{\gamma} = \bm{0}$, so the equations \eqref{eq:lipm} and \eqref{eq:dcm-dynamics} are equivalent to the conventional equations when there are no external forces.

\section{Pattern Generator with External Forces} \label{sec:pg}

The pattern generator (PG) generates a CoM trajectory to track the input reference ZMP trajectory.
The reference ZMP trajectory is determined by the footstep sequence obtained from the operator command or planner~\cite{LocomanipPlan:Murooka:RAL2021} at the upper layer.
In loco-manipulation, the reference trajectories of the force, moment, and position of manipulation external contacts (e.g., hands) are also inputted to the PG.
In this letter, we present a PG that extends the preview control in~\cite{PreviewControl:Kajita:ICRA2003}.
See the end of this section for other methods extensions.

\subsection{Introduction of ext-ZMP}

LIPM with external forces \eqref{eq:lipm} can be expressed in the same format as the conventional LIPM by replacing a variable as follows:
\begin{align}
  & \bm{\ddot{c}} = \omega^2 \left( \bm{c} - \bm{\hat{z}} \right) \label{eq:lipm-ex-zmp}
  & {\rm where} \ \ \ \ \bm{\hat{z}} = \kappa \bm{z} - \bm{\gamma}
\end{align}
In $\bm{\hat{z}}$, the scale and offset due to external forces are added to $\bm{z}$.
$\bm{\hat{z}}$ is afterward referred to as ZMP with external forces (ext-ZMP).
If the external forces are zero, ext-ZMP coincides with the conventional ZMP, making it a straightforward extension.

\subsection{Preview Control}

By replacing the conventional ZMP with an ext-ZMP, the conventional preview control can be applied as it is.
The following derivation of the preview control just follows~\cite{PreviewControl:Kajita:ICRA2003}, except that ZMP is replaced by ext-ZMP, but we describe it for self-contained explanation.

In the following, we focus only on the X component as the Y component can be handled in the same way.
By discretizing \eqref{eq:lipm-ex-zmp} and considering the CoM jerk as the input and the ZMP as the output, the following state equation is obtained:
\begin{subequations}
\label{eq:pg-state-eq}
\begin{align}
  \bm{x_c}[k+1]
  &=
  \begin{bmatrix} 1 & \Delta t & \Delta t^2 / 2 \\ 0 & 1 & \Delta t \\ 0 & 0 & 1 \end{bmatrix}
  \! \bm{x_c}[k]
  \!+\!
  \begin{bmatrix} \Delta t^3 / 6 \\ \Delta t^2 / 2 \\ \Delta t \end{bmatrix}
  \! u_x[k] \\
  \hat{z}_x[k] &= \begin{bmatrix} 1 & 0 & - 1 / \omega^2 \end{bmatrix}
  \bm{x_c}[k] \\
  & {\rm where} \ \ \ \ \bm{x_c}[k] = \begin{bmatrix} c_x[k] & \dot{c}_x[k] & \ddot{c}_x[k] \end{bmatrix}^T \nonumber
\end{align}
\end{subequations}
$\Delta t$ is the control time-step and $k$ is the step index.
By minimizing the following objective function, the CoM trajectory that tracks the reference ext-ZMP (i.e., tracks the reference ZMP and the reference external forces) can be obtained:
\begin{align}
  J = \sum_{j=k}^{\infty} \left( Q \left( \hat{z}_x[j] - \hat{z}_x^{\mathit{ref}}[j] \right)^2 + R \, u_x^2[j] \right) \label{eq:pg-objective}
\end{align}
$Q$ and $R$ are the objective weights.
The control input (CoM jerk) that minimizes \eqref{eq:pg-objective} under \eqref{eq:pg-state-eq} is obtained by the following equation~\cite{PreviewControl:Kajita:ICRA2003}:
\begin{align}
  u_x[k] = - \bm{K}_{\mathit{fb}} \, \bm{x_c}[k] + \sum_{j=1}^{N_h} K_{\mathit{ff}}[j] \, \hat{z}_x^{\mathit{ref}}[k+j]
\end{align}
$N_h$ is the number of time-steps of the preview window.
The control gains $\bm{K}_{\mathit{fb}}$ and $K_{\mathit{ff}}$ are determined from the system matrices in \eqref{eq:pg-state-eq} and objective weights in \eqref{eq:pg-objective} \cite{PreviewControl:Katayama:IJC1985}.
Note that the effect of the external forces is isolated to ext-ZMP, and the control gains remain constant even when the external forces change.

Although there are some previous studies that use preview control for the motion of pushing an object~\cite{PushZMP:Harada:ICRA2003,Push:Takubo:ICRA2005,Push:Nishiwaki:Humanoids2006,Push:Stilman:ICRA2008,Push:Hakamata:AR2020}, they only add an offset to the reference ZMP or the output CoM, which is equivalent to setting the ZMP scale $\kappa$ to 1 in ext-ZMP \eqref{eq:lipm-ex-zmp}. In this letter, the ZMP scale due to the vertical external forces is correctly considered, and as shown in Section~\ref{sec:val}, the walking stability is improved when the robot is subject to vertical external forces.

\subsection{Other Pattern Generation Methods}
The formulas in Section~\ref{sec:basic} do not prohibit that other pattern generation methods may also be easily extended to treat external manipulation forces. For example, in the linear MPC-based method~\cite{MPCWalk:Herdt:AR2010}, the state equation \eqref{eq:pg-state-eq} and the objective function \eqref{eq:pg-objective} can be used as they are.
In addition, DCM-based method~\cite{DcmWalk:Englsberger:TRO2015} can make use of the extension of the DCM dynamics shown in~\eqref{eq:dcm-dynamics}. Implementing these extensions on our robots requires additional engineering efforts that are beyond the scope of this letter.

\section{Stabilizer with External Forces} \label{sec:st}

\begin{figure}[tpb]
  \begin{center}
    \includegraphics[width=1.0\columnwidth]{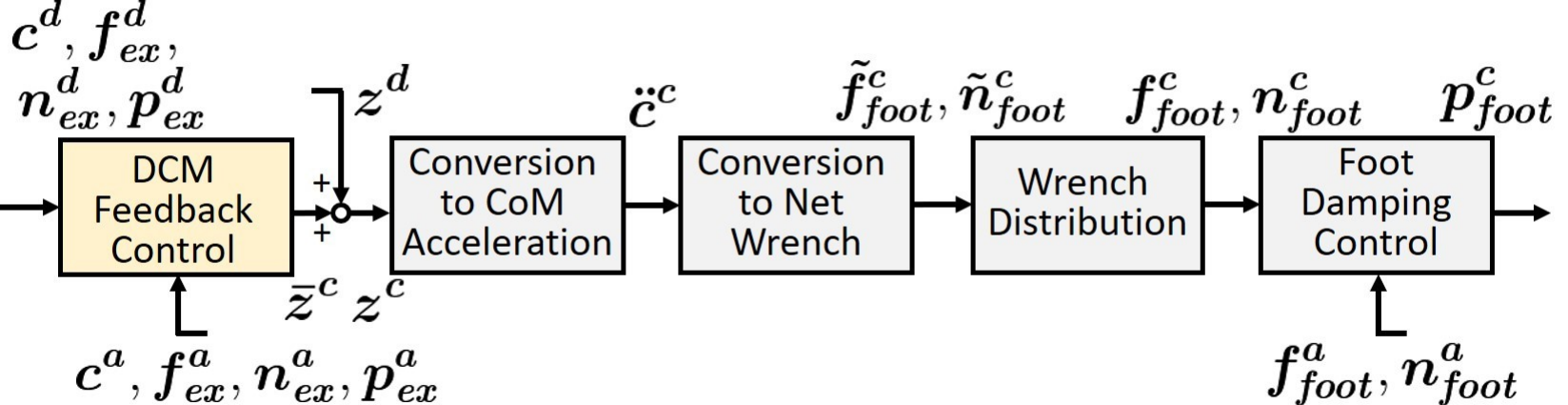}
    \caption{Calculation procedure of the stabilizer.
      \newline \footnotesize
      ``DCM Feedback Control'' is extended to treat external manipulation forces.
    }
    \label{fig:st}
  \end{center}
\end{figure}

In the stabilizer (ST), the error between the CoM obtained from PG and the actual estimated robot CoM is reduced based on sensor measurements.
In this letter, we employ an ST based on DCM feedback control~\cite{StabilizerIntegral:Morisawa:Humanoids2012,DcmWalk:Englsberger:TRO2015,StairClimb:Caron:ICRA2019}.
Particularly, we extend the implementation of~\cite{StairClimb:Caron:ICRA2019}, whose source code is open.
\figref{fig:st} shows the calculation procedure in the ST~\cite{StairClimb:Caron:ICRA2019}.
In this procedure, ``DCM Feedback Control'' is extended to account for external manipulation forces.

\subsection{Strategy Overview} \label{sec:st-strategy}

\begin{figure}[tpb]
  \begin{center}
    \includegraphics[width=1.0\columnwidth]{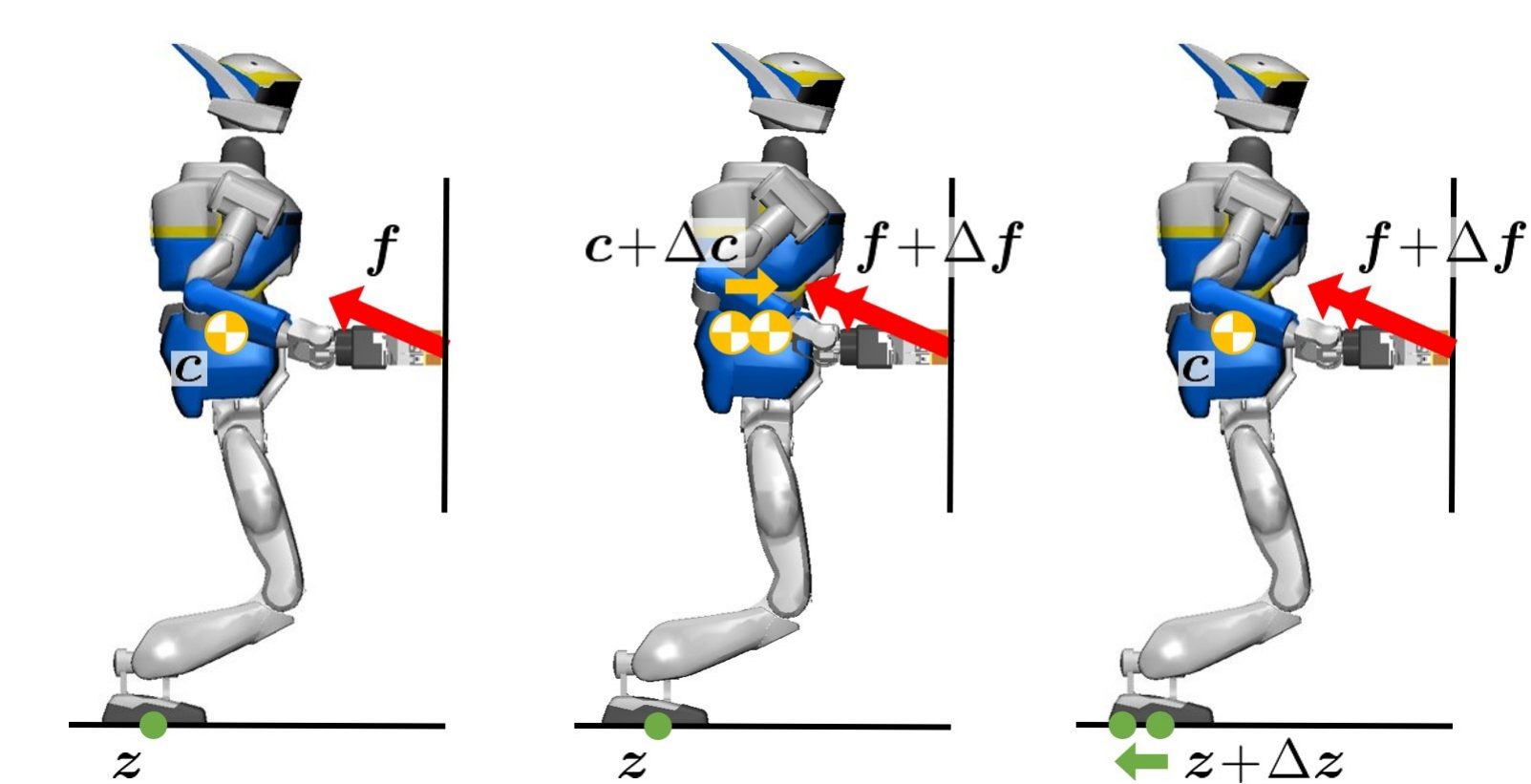}\\
    \begin{minipage}{0.32\columnwidth}
      \begin{center} \footnotesize Nominal state \end{center}
    \end{minipage}
    \begin{minipage}{0.32\columnwidth}
      \begin{center} \footnotesize CoM strategy \end{center}
    \end{minipage}
    \begin{minipage}{0.32\columnwidth}
      \begin{center} \footnotesize ZMP strategy \end{center}
    \end{minipage}
    \caption{Strategy for dealing with external force error in stabilizer.
      \newline \footnotesize
      In the example shown in the figure, the robot is pushed backward by forces larger than expected at hands.
      In the CoM strategy, the robot moves the CoM forward while maintaining the ZMP. In the ZMP strategy, the robot moves the ZMP backwards while maintaining the CoM.
    }
    \label{fig:st-strategy}
  \end{center}
\end{figure}

The proposed ST uses two strategies to maintain balance in response to external manipulation forces error: CoM strategy and ZMP strategy (\figref{fig:st-strategy}).
CoM strategy can handle larger errors than ZMP strategy, which is constrained by the sole region.
Conversely, the ZMP strategy can respond to errors faster compared to the CoM strategy. The latter requires whole-body robot motions.
The proposed ST separates the errors in the frequency domain and applies these strategies in a complementary manner.

\subsection{DCM Feedback Control} \label{sec:st-dcm}

First, the state equation of the DCM error including the error of the external forces is derived in Sections~\ref{sec:st-dcm-system-1}-5, and then the feedback law is introduced in Section~\ref{sec:st-dcm-control}.

\subsubsection{Dynamics of DCM Error} \label{sec:st-dcm-system-1}

From \eqref{eq:dcm-dynamics}, the following equations hold for the actual value $\bm{*^a}$ obtained from sensor measurements and the desired value $\bm{*^d}$ obtained from the loco-manipulation PG:
\begin{subequations}
\begin{align}
  \bm{\dot{\xi}^a} &= \omega (\bm{\xi^a} - \kappa^a \bm{z^a} + \bm{\gamma^a}) \label{eq:dcm-dynamics-a} \\
  \bm{\dot{\xi}^d} &= \omega (\bm{\xi^d} - \kappa^d \bm{z^d} + \bm{\gamma^d}) \label{eq:dcm-dynamics-d}
\end{align}
\end{subequations}
To linearize the error dynamics, we assume that $\kappa = \kappa^a = \kappa^d$ is constant.
In general motion, the error of $\kappa$ in \eqref{eq:lipm-scale} due to the error of the vertical manipulation (hand) forces is not so large, thus this simplification does not have much impact.

Subtracting both sides of \eqref{eq:dcm-dynamics-d} from \eqref{eq:dcm-dynamics-a}, we have:
\begin{align}
  \bm{\dot{\bar{\xi}}^a} &= \omega (\bm{\bar{\xi}^a} - \kappa \bm{\bar{z}^a} + \bm{\bar{\gamma}^a}) \label{eq:dcm-dynamics-err}
\end{align}
$\bm{\bar{*}^a}$ denotes the error between the actual value and desired value ($\bm{*^a} - \bm{*^d}$).

\subsubsection{Dynamics of ZMP Delay}

In the ST, the DCM is controlled by realizing the command ZMP obtained through feedback with the foot damping control~\cite{Stabilizer:Kajita:IROS2010,StairClimb:Caron:ICRA2019}.
In reality, the actual ZMP lags behind the command ZMP due to mechanical compliance and control tracking.
It was proposed to deal with this delay with the following first-order lag system \cite{Stabilizer:Kajita:IROS2010,StabilizerIntegral:Morisawa:Humanoids2012}:
\begin{align}
  \bm{\dot{\bar{z}}^a} = - \rho \bm{\bar{z}^a} + \rho \bm{\bar{z}^c} \label{eq:zmp-delay}
\end{align}
$\rho$ is the parameter of the first-order lag system.
$\bm{\bar{z}^c} = \bm{z^c} - \bm{z^d}$ is the difference between command ZMP and desired ZMP.

\subsubsection{Second-order Dynamics of DCM Error}

Differentiating \eqref{eq:dcm-dynamics-err} leads to:
\begin{subequations}
\begin{align}
  \bm{\ddot{\bar{\xi}}^a} &= \omega \left( \bm{\dot{\bar{\xi}}^a} - \kappa \bm{\dot{\bar{z}}^a} + \bm{\dot{\bar{\gamma}}^a} \right) \label{eq:dcm-dynamics-second-1} \\
  &= \omega \left( \bm{\dot{\bar{\xi}}^a} - \kappa \left( - \rho \bm{\bar{z}^a} + \rho \bm{\bar{z}^c} \right) + \bm{\dot{\bar{\gamma}}^a} \right) \label{eq:dcm-dynamics-second-2} \\
  &= \rho \omega \bm{\bar{\xi}^a} + \left( \omega - \rho \right) \bm{\dot{\bar{\xi}}^a} - \kappa \rho \omega \bm{\bar{z}^c} + \rho \omega \bm{\bar{\gamma}^a} + \omega \bm{\dot{\bar{\gamma}}^a} \label{eq:dcm-dynamics-second}
\end{align}
\end{subequations}
\eqref{eq:dcm-dynamics-second} is the second-order dynamics of the DCM error used to derive the feedback law of ST.
\eqref{eq:zmp-delay} is used for the transformation from \eqref{eq:dcm-dynamics-second-1} to \eqref{eq:dcm-dynamics-second-2}, and \eqref{eq:dcm-dynamics-err} is used for the transformation from \eqref{eq:dcm-dynamics-second-2} to \eqref{eq:dcm-dynamics-second}.

\subsubsection{Frequency-Domain Separation of External Forces}

For the complementary implementation of the CoM strategy and ZMP strategy described in Section~\ref{sec:st-strategy}, the offset $\bm{\bar{\gamma}^a}$ due to external forces is separated into high-frequency component $\bm{\bar{\gamma}^a_H}$ and low-frequency component $\bm{\bar{\gamma}^a_L}$:
\begin{align}
  \bm{\bar{\gamma}^a} = \bm{\bar{\gamma}^a_H} + \bm{\bar{\gamma}^a_L}
\end{align}
The low-frequency component is handled by the CoM strategy by replacing the desired CoM with the following $\bm{c^{\prime d}}$ instead of $\bm{c^d}$ as follows:
\begin{align}
  \bm{c^{\prime d}} = \bm{c^d} - \bm{\bar{\gamma}^a_L}
\end{align}
Assuming that the rate of change is negligible in the low-frequency component ($\bm{\dot{\bar{\gamma}}^a_L} \approx 0$), the desired DCM and the DCM error are expressed as:
\begin{subequations}
\begin{align}
  \bm{\xi^{\prime d}} &= \bm{c^{\prime d}} + \frac{1}{\omega} \bm{\dot{c}^{\prime d}}
  = \bm{\xi^d} - \bm{\bar{\gamma}^a_L} \\
  \bm{\bar{\xi}^{\prime a}} &= \bm{\xi^a} - \bm{\xi^{\prime d}}
  = \bm{\bar{\xi}^a} + \bm{\bar{\gamma}^a_L} \label{eq:xi-err-offset}
\end{align}
\end{subequations}
Substituting \eqref{eq:xi-err-offset} into \eqref{eq:dcm-dynamics-second} cancels $\bm{\bar{\gamma}^a_L}$ and gives:
\begin{align}
  \bm{\ddot{\bar{\xi}}^{\prime a}}
  &= \rho \omega \bm{\bar{\xi}^{\prime a}} + \left( \omega - \rho \right) \bm{\dot{\bar{\xi}}^{\prime a}} - \kappa \rho \omega \bm{\bar{z}^c} + \rho \omega \bm{\bar{\gamma}^a_H} + \omega \bm{\dot{\bar{\gamma}}^a_H} \label{eq:dcm-dynamics-second-freq}
\end{align}

\subsubsection{State Equation}

In the following, we reason on the X component as the Y component can be handled in the same way.
\eqref{eq:dcm-dynamics-second-freq} is expressed by the following state equation:
\begin{align}
  &\frac{d}{dt} \bm{x_\xi} =
  \bm{A_\xi} \bm{x_\xi} + \bm{B_\xi} \bar{z}^c_x + \bm{C_\xi} \label{eq:st-state-eq} \\
  & {\rm where} \ \
  \bm{x_\xi} = \begin{bmatrix} \int \! \bar{\xi}^{\prime a}_x dt \\ \bar{\xi}^{\prime a}_x \\ \dot{\bar{\xi}}^{\prime a}_x \end{bmatrix}, \ \
  \bm{A_\xi} = \begin{bmatrix} 0 & 1 & 0 \\ 0 & 0 & 1 \\ 0 & \rho \omega & \omega \!-\! \rho \end{bmatrix} \nonumber \\
  & \phantom{{\rm where} \ \ }
  \bm{B_\xi} = \begin{bmatrix} 0 \\ 0 \\ - \kappa \rho \omega \end{bmatrix}, \ \
  \bm{C_\xi} = \begin{bmatrix} 0 \\ 0 \\ \rho \omega \bar{\gamma}^a_{H,x} + \omega \dot{\bar{\gamma}}^a_{H,x} \end{bmatrix} \nonumber
\end{align}
The system matrix $\bm{A_\xi}$ does not depend on external forces.

\subsubsection{Feedback Control} \label{sec:st-dcm-control}

We construct the following state feedback for the system \eqref{eq:st-state-eq}:
\begin{align}
  \bar{z}^c_x &= \begin{bmatrix} \tilde{k}_i & \tilde{k}_p & \tilde{k}_d \end{bmatrix} \bm{x_\xi}
  + \frac{1}{\kappa} \bar{\gamma}^a_{H,x} + \frac{1}{\kappa \rho} \dot{\bar{\gamma}}^a_{H,x} \label{eq:dcm-feedback}
\end{align}
This is the PID control of DCM via ZMP~\cite{StabilizerIntegral:Morisawa:Humanoids2012,StairClimb:Caron:ICRA2019}.
Then, substituting \eqref{eq:dcm-feedback} into \eqref{eq:st-state-eq}, the following closed-loop response is obtained:
\begin{align}
  \frac{d}{dt} \bm{x_\xi} =
  \begin{bmatrix} 0 \!&\! 1 \!&\! 0 \\ 0 \!&\! 0 \!&\! 1 \\ - \kappa \rho \omega \tilde{k}_i \!&\! \rho \omega (1 \!-\! \kappa \tilde{k}_p) \!&\! \omega \!-\! \rho \!-\! \kappa \rho \omega \tilde{k}_d \end{bmatrix} \bm{x_\xi} \label{eq:dcm-closed-loop}
\end{align}
$\bm{\bar{\gamma}^a_H}$ is canceled by the last two terms of \eqref{eq:dcm-feedback}.
The closed-loop response \eqref{eq:dcm-closed-loop} does not depend on $\bm{\gamma}$, but only on $\kappa$.
The system is stabilized by determining the gains $\tilde{k}_p, \tilde{k}_i, \tilde{k}_d$ by pole placement~\cite{StabilizerIntegral:Morisawa:Humanoids2012} or experimentally~\cite{StairClimb:Caron:ICRA2019}.

The closed-loop response of the conventional ST without external forces under the same feedback law is as follows:
\begin{align}
  \frac{d}{dt} \bm{x_\xi} =
  \begin{bmatrix} 0 \!&\! 1 \!&\! 0 \\ 0 \!&\! 0 \!&\! 1 \\ - \rho \omega {k}_i \!&\! \rho \omega (1 \!-\! {k}_p) \!&\! \omega \!-\! \rho \!-\! \rho \omega {k}_d \end{bmatrix} \bm{x_\xi} \label{eq:dcm-closed-loop-conventional}
\end{align}
Comparing \eqref{eq:dcm-closed-loop} and \eqref{eq:dcm-closed-loop-conventional}, to match the response of the proposed ST with the response of the conventional ST of the gains $k_p, k_i, k_d$, the gains in \eqref{eq:dcm-feedback} should be set as follows:
\begin{align}
  \tilde{k}_p = \frac{k_p}{\kappa}, \ \
  \tilde{k}_i = \frac{k_i}{\kappa}, \ \
  \tilde{k}_d = \frac{k_d}{\kappa}
\end{align}

\begin{figure}[tpb]
  \begin{center}
    \includegraphics[width=0.9\columnwidth]{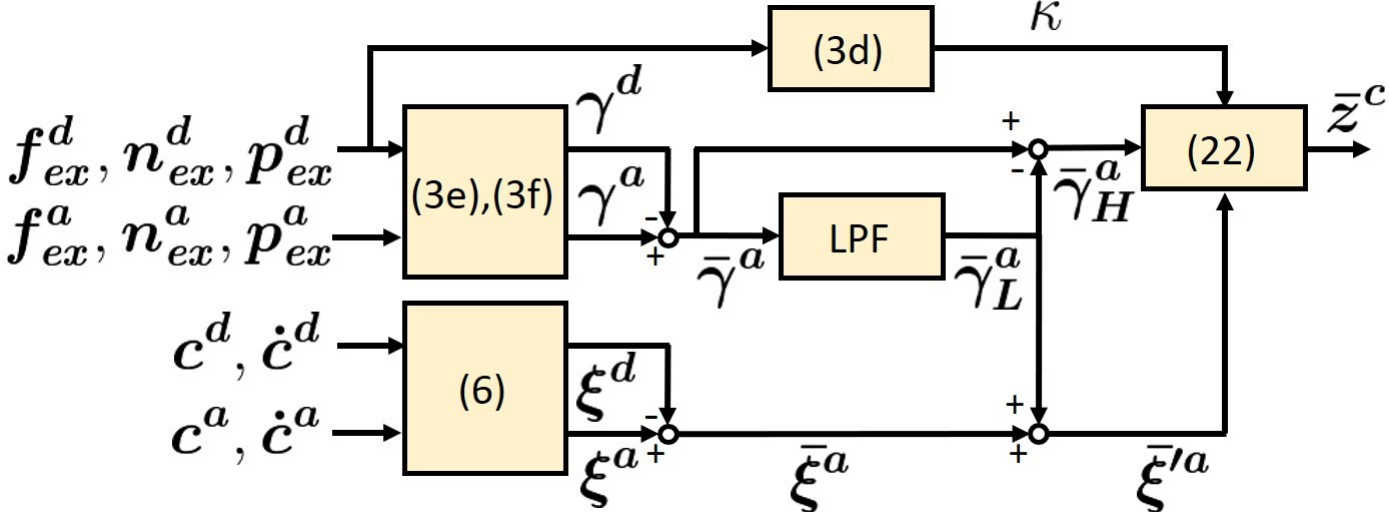}
    \caption{Calculation procedure of DCM feedback control.
    }
    \label{fig:st-dcm}
  \end{center}
\end{figure}

\figref{fig:st-dcm} shows the calculation procedure of DCM feedback control considering external forces.

\subsection{Overall Process of Stabilizer}

The control ZMP $\bm{\bar{z}^c} = [ \bar{z}^c_x \ \bar{z}^c_y]^T$ calculated by \eqref{eq:dcm-feedback} is converted according to the ST calculation procedure (\figref{fig:st}).
The rest of the processes only follow~\cite{StairClimb:Caron:ICRA2019}, so it is omitted as appropriate.

From \eqref{eq:dcm-feedback}, the command ZMP $\bm{z^c}$ can be obtained as:
\begin{align}
  \bm{z^c} &= \bm{z^d} + \frac{1}{\kappa} \mathit{PID} + \frac{1}{\kappa} \bm{\bar{\gamma}^a_{H}} + \frac{1}{\kappa \rho} \bm{\dot{\bar{\gamma}}^a_{H}} \label{eq:command-zmp} \\
  &{\rm where} \ \ \mathit{PID} = k_p \bm{\bar{\xi}^{\prime a}} + k_i \int \! \bm{\bar{\xi}^{\prime a}} dt + k_d \bm{\dot{\bar{\xi}}^{\prime a}} \nonumber
\end{align}
From \eqref{eq:lipm} and \eqref{eq:command-zmp}, the command ZMP $\bm{z^c}$ is converted to the command CoM acceleration $\bm{\ddot{c}^c}$:
\begin{subequations}
\begin{align}
  \bm{\ddot{c}^c} &= \omega^2 (\bm{c^d} - \kappa \bm{z^c} + \bm{\gamma^d}) \\
  &= \bm{\ddot{c}^d} - \omega^2 \mathit{PID} - \omega^2 \bm{\bar{\gamma}^a_{H}} - \frac{\omega^2}{\rho} \bm{\dot{\bar{\gamma}}^a_{H}} \label{eq:command-com-accel}
\end{align}
\end{subequations}
From \eqref{eq:newton-euler} and \eqref{eq:command-com-accel}, the command CoM acceleration $\bm{\ddot{c}^c}$ is converted to the command net force and moment of feet represented in the world frame $\bm{\tilde{f}^c_{\mathit{foot}}}, \bm{\tilde{n}^c_{\mathit{foot}}}$:
\begin{align}
  \begin{bmatrix} \bm{\tilde{f}^c_{\mathit{foot}}} \\ \bm{\tilde{n}^c_{\mathit{foot}}} \end{bmatrix}
  =
  \begin{bmatrix} m ( \bm{\ddot{c}^c} + \bm{g} ) \\ \bm{c^c} \times \bm{\tilde{f}^c_{\mathit{foot}}} \end{bmatrix}
  -
  \sum_i
  \begin{bmatrix}
    \bm{f^{[i]}_{\mathit{ex}}} \\
    \left(\bm{p^{[i]}_{\mathit{ex}}} - \bm{c}\right) \times \bm{f^{[i]}_{\mathit{ex}}} + \bm{n^{[i]}_{\mathit{ex}}}
  \end{bmatrix}
  \label{eq:command-net-wrench}
\end{align}
Here, the angular momentum rate around the world origin is assumed to be zero.
The command force and moment for each foot $\bm{f^c_{\mathit{foot}}}, \bm{n^c_{\mathit{foot}}}$ are obtained by solving the quadratic programming (QP) that distributes the net force and moment $\bm{\tilde{f}^c_{\mathit{foot}}}, \bm{\tilde{n}^c_{\mathit{foot}}}$ to the left and right feet with saturation to the sole region~\cite{StairClimb:Caron:ICRA2019}.
Finally, the command force and moment are realized by foot damping control~\cite{Stabilizer:Kajita:IROS2010,StairClimb:Caron:ICRA2019}.

\section{Implementation and Simulation} \label{sec:val}

The proposed PG and ST are implemented in {\tt C++} within a real-time robot control framework {\tt mc\_rtc}~\cite{mc_rtc:github2021}, along with impedance control of the hand~\cite{Push:Stilman:ICRA2008,Push:Nishiwaki:Humanoids2006,OpenDoor:Nozawa:IROS2012} and the tracking control of the swing foot.
Kinematics commands, such as the hand, foot, and CoM positions, are passed to the acceleration-based whole-body inverse kinematics calculation, and the calculated joint angles are commanded to the low-level robot's position PD controller.
The proposed PG and ST are computationally inexpensive and can be executed in 2~ms cycles on a robot's embedded computer.
We used the following values as parameters for PG and ST: $Q = 1, R = 10^{-8}$ in \eqref{eq:pg-objective}, $\tilde{k}_i = 0, \tilde{k}_p = 1.25, \tilde{k}_d = 0$ in \eqref{eq:dcm-feedback}.


The proposed PG and ST are assessed with three test-cases with the dynamics simulator Choreonoid~\cite{Choreonoid:Nakaoka:SII2012}.
In the verifications, we used a simulation function that applies specified external forces to the robot's hands.

\begin{figure*}[t]
  \minipage{0.195\textwidth}
  \includegraphics[width=\linewidth]{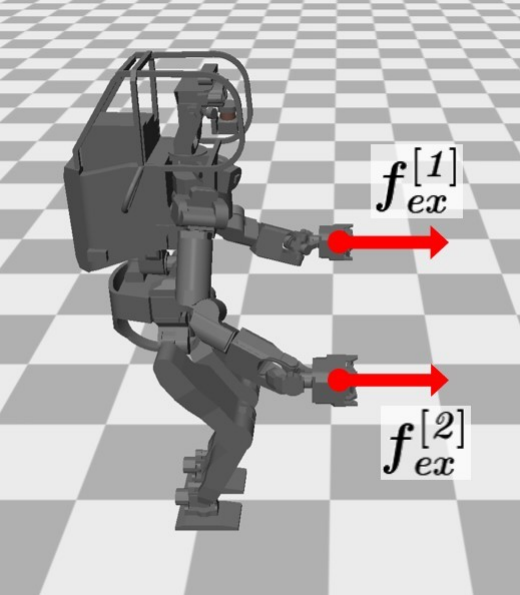}
  \caption{Test-cases 1 and 3.}
  \label{fig:val1}
  \endminipage\hfill
  \minipage{0.78\textwidth}
  \includegraphics[width=\linewidth, clip, bb=155 4 1691 421]{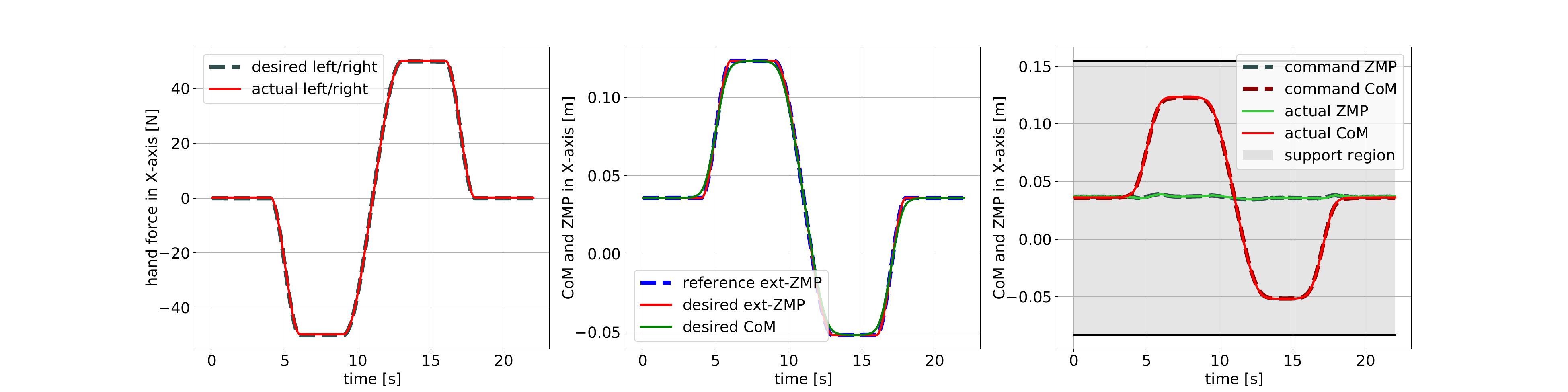}\\
  \begin{minipage}{0.32\linewidth}
    \begin{center} \footnotesize \hspace{6mm} (A) Hand forces \end{center}
  \end{minipage}
  \begin{minipage}{0.32\linewidth}
    \begin{center} \footnotesize \hspace{6mm} (B) PG input and output \end{center}
  \end{minipage}
  \begin{minipage}{0.32\linewidth}
    \begin{center} \footnotesize \hspace{6mm} (C) CoM and ZMP \end{center}
  \end{minipage}
  \vspace{-1.5mm}
  \caption{Results of test-case~1, where the PG generates trajectories considering the external forces on the X-axis.}
  \label{fig:val-pg-x}
  \endminipage
  \vspace{4mm}
  \minipage{0.195\textwidth}
  \includegraphics[width=\linewidth]{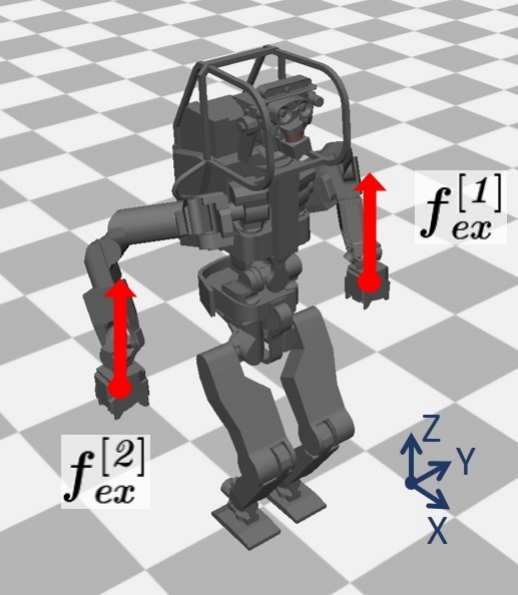}
  \caption{Test-case 2.}
  \label{fig:val2}
  \endminipage\hfill
  \minipage{0.78\textwidth}
  \includegraphics[width=\linewidth, clip, bb=155 4 1691 421]{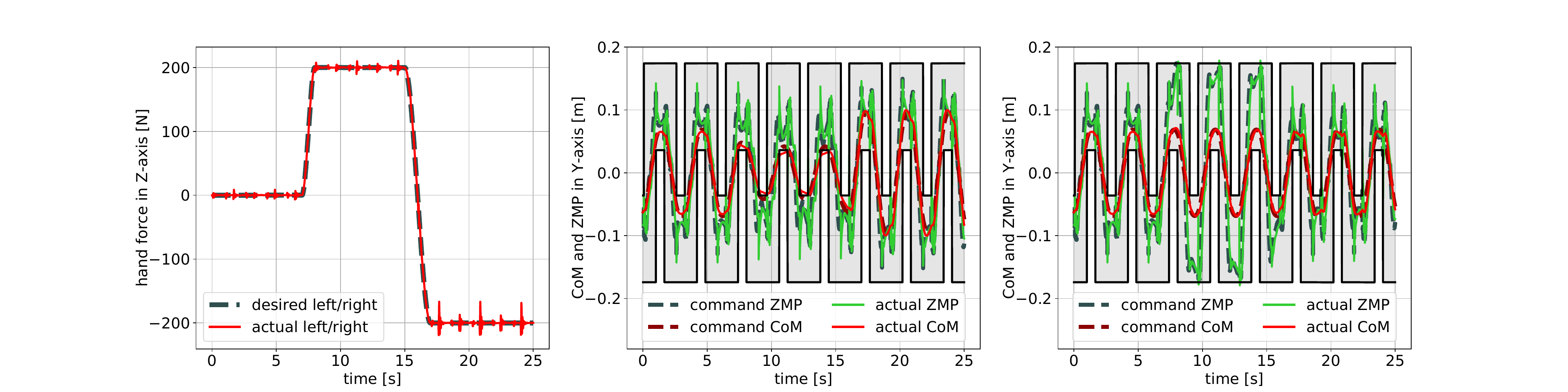}\\
  \begin{minipage}{0.32\linewidth}
    \begin{center} \footnotesize \hspace{6mm} (A) Hand forces \end{center}
  \end{minipage}
  \begin{minipage}{0.32\linewidth}
    \begin{center} \footnotesize \hspace{6mm} (B) CoM and ZMP \end{center}
  \end{minipage}
  \begin{minipage}{0.32\linewidth}
    \begin{center} \footnotesize \hspace{6mm} (C) Without the proposed PG \end{center}
  \end{minipage}
  \vspace{-1.5mm}
  \caption{Results of test-case~2, where the PG generates trajectories considering the external forces on the Z-axis.}
  \label{fig:val-pg-z}
  \endminipage
  \vspace{4mm}
  \includegraphics[width=\linewidth, clip, bb=130 4 2200 419]{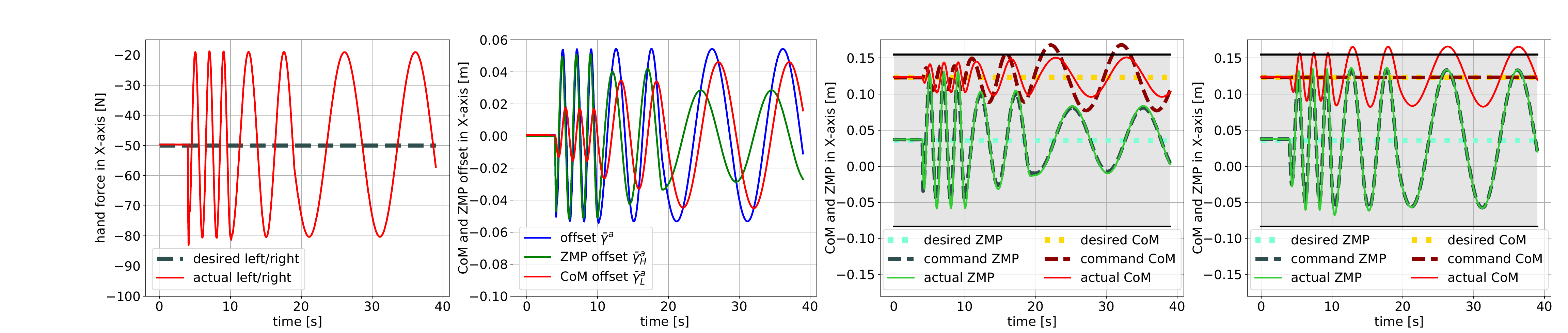}\\
  \begin{minipage}{0.22\linewidth}
    \begin{center} \footnotesize \hspace{6mm} (A) Hand forces \end{center}
  \end{minipage}
  \begin{minipage}{0.26\linewidth}
    \begin{center} \footnotesize \hspace{6mm} (B) Offsets due to external forces \end{center}
  \end{minipage}
  \begin{minipage}{0.25\linewidth}
    \begin{center} \footnotesize \hspace{6mm} (C) CoM and ZMP \end{center}
  \end{minipage}
  \begin{minipage}{0.23\linewidth}
    \begin{center} \footnotesize \hspace{6mm} (D) Without the proposed ST \end{center}
  \end{minipage}
  \vspace{-1.5mm}
  \caption{Results of test-case~3, where the ST compensates for the external force disturbances on the X-axis.}
  \label{fig:val-st-disturb}
\end{figure*}

In test-case~1, we verify that the PG generates a CoM trajectory consistent with the given reference trajectories of horizontal external forces (\figref{fig:val1}).
\figref{fig:val-pg-x} shows the results.
As shown in (A), forces in the front-back direction of 50~N are applied to each hand (100~N in total on the left and right).
(B) shows that the PG generates a CoM trajectory that tracks the ext-ZMP calculated from the external forces, and (C) shows that the robot successfully responds to the external forces with almost no fluctuation in the ZMP.
Even if the robot is constantly stepping on the spot, obtained result is similar to static standing.

Test-case~2 verifies the response to external forces in the vertical direction (\figref{fig:val2}).
\figref{fig:val-pg-z} shows the results.
As shown in (A), upward and downward forces of 200~N are applied to each hand while the robot fixes the hand position and steps on the spot.
(B) shows that the robot is stepping while keeping the ZMP in the center of the support region.
Note that the PG generates a CoM trajectory with different amounts of lateral sway depending on the vertical external forces.
This is because, as mentioned in Section~\ref{sec:pg}, the PG correctly considers the ZMP scale effect due to the vertical external forces.
As shown in (C), ignoring this effect (setting $\kappa = 1$ in \eqref{eq:lipm-ex-zmp}) will cause ZMP to fluctuate due to external forces.

In test-case~3, while applying the same 50~N backward force as in test-case~1, we exert on each hand a sinusoidal disturbance force in the same axis (X-axis), with an amplitude of 30~N.
\figref{fig:val-st-disturb} shows the results.
As shown in (A), disturbances with periods of 2~s, 5~s, and 10~s were applied in sequence.
(B) shows the offset $\bm{\bar{\gamma}^a}$ due to disturbances, the high-frequency component $\bm{\bar{\gamma}^a_{H}}$ handled by the ZMP strategy, and the low-frequency component $\bm{\bar{\gamma}^a_{L}}$ handled by the CoM strategy.
The cutoff period of the low-pass filter for separation is 1.0~s.
As shown in (C), high-frequency disturbances are dealt with by adjusting ZMP, and low-frequency disturbances are dealt with by adjusting CoM.
As shown in (D), when the compensation for external force error in ST is disabled ($\bm{\bar{\gamma}^a_{H}} = \bm{\bar{\gamma}^a_{L}} = \bm{0}$), the fluctuations of CoM and ZMP become large, which indicates the effectiveness of the external force error compensation in the proposed ST.

\section{Application to Loco-manipulation Tasks} \label{sec:exp}

We applied the proposed controller to loco-manipulation motions with a humanoid robot in simulation and the real world.

\begin{figure*}[t]
  \includegraphics[width=0.23\linewidth]{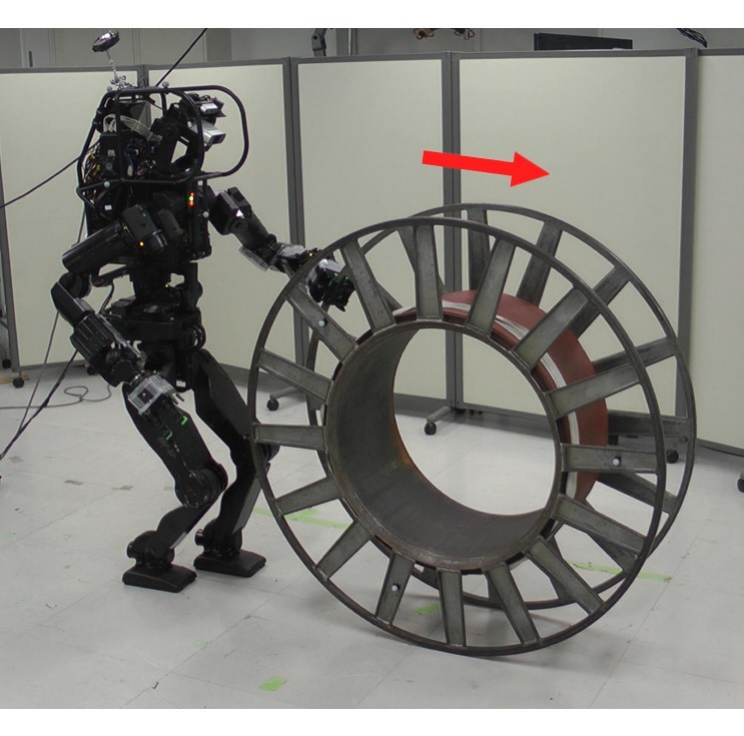}
  \includegraphics[width=0.23\linewidth]{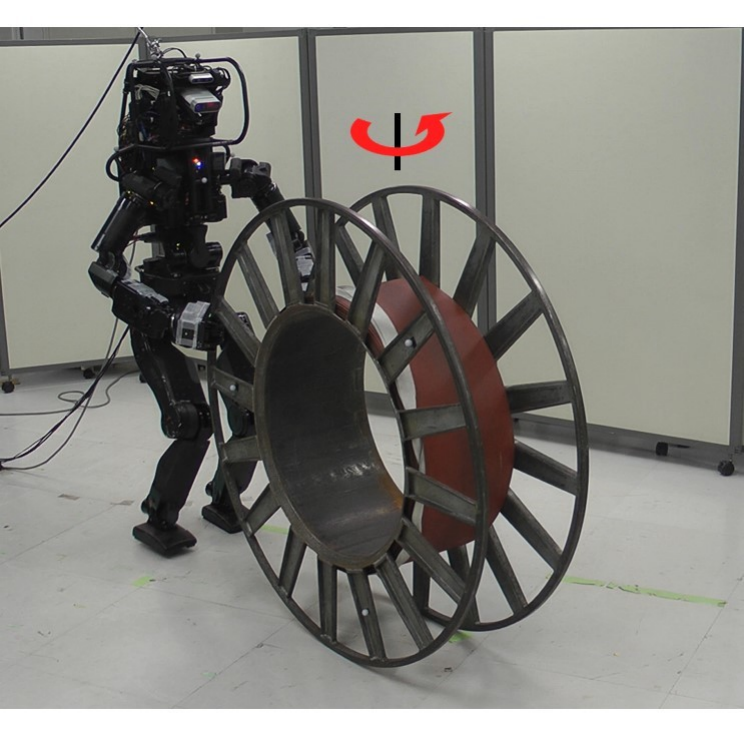}
  \hfill
  \includegraphics[width=0.52\linewidth, clip, bb=77 4 1066 421]{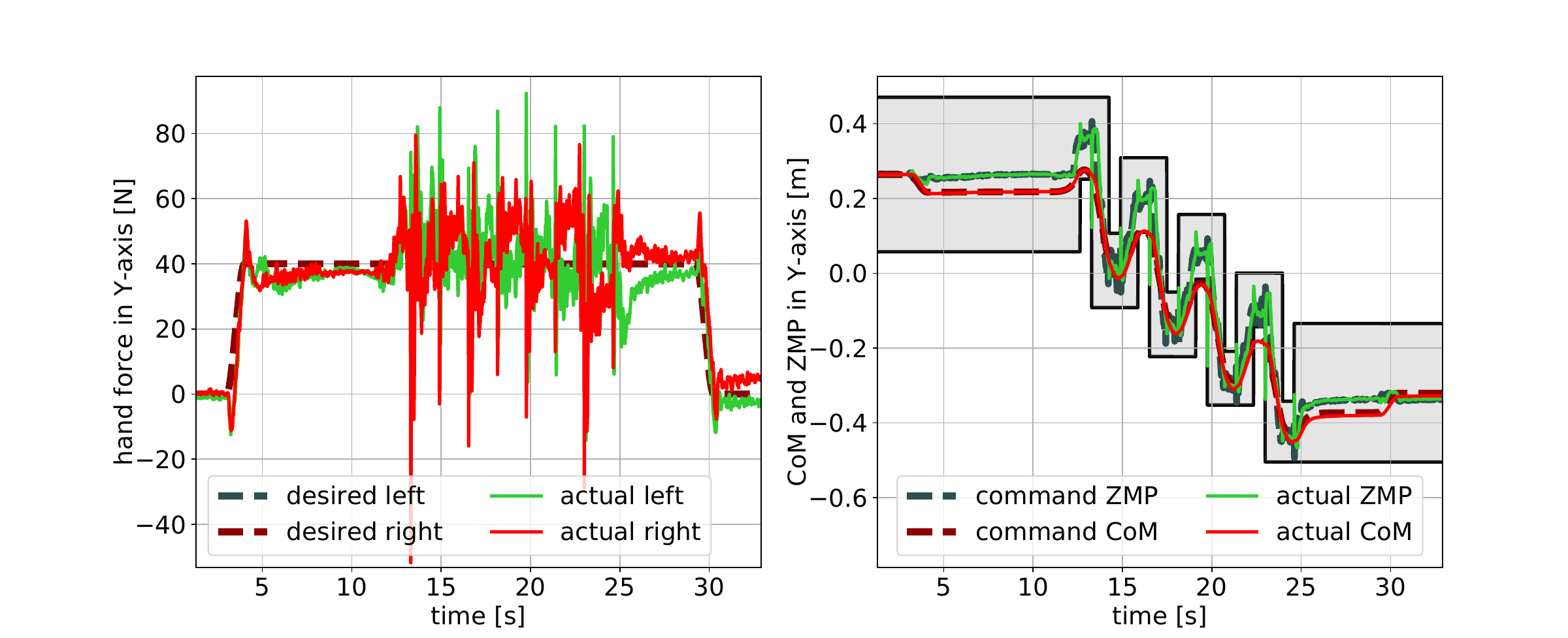}\\
  \begin{minipage}{0.22\linewidth}
    \begin{center} \footnotesize \hspace{2mm} (A) Forward motion \end{center}
  \end{minipage}
  \begin{minipage}{0.22\linewidth}
    \begin{center} \footnotesize \hspace{2mm} (B) Turning motion \end{center}
  \end{minipage}
  \hfill
  \begin{minipage}{0.26\linewidth}
    \begin{center} \footnotesize \hspace{6mm} (C) Hand forces during turning \end{center}
  \end{minipage}
  \begin{minipage}{0.26\linewidth}
    \begin{center} \footnotesize \hspace{6mm} (D) CoM and ZMP during turning \end{center}
  \end{minipage}
  \vspace{-1.5mm}
  \caption{Bobbin rolling operation by HRP-5P.
  }
  \label{fig:exp-bobbin}
  \vspace{3mm}
  \includegraphics[width=1.0\linewidth, clip, bb=16 4 1851 421]{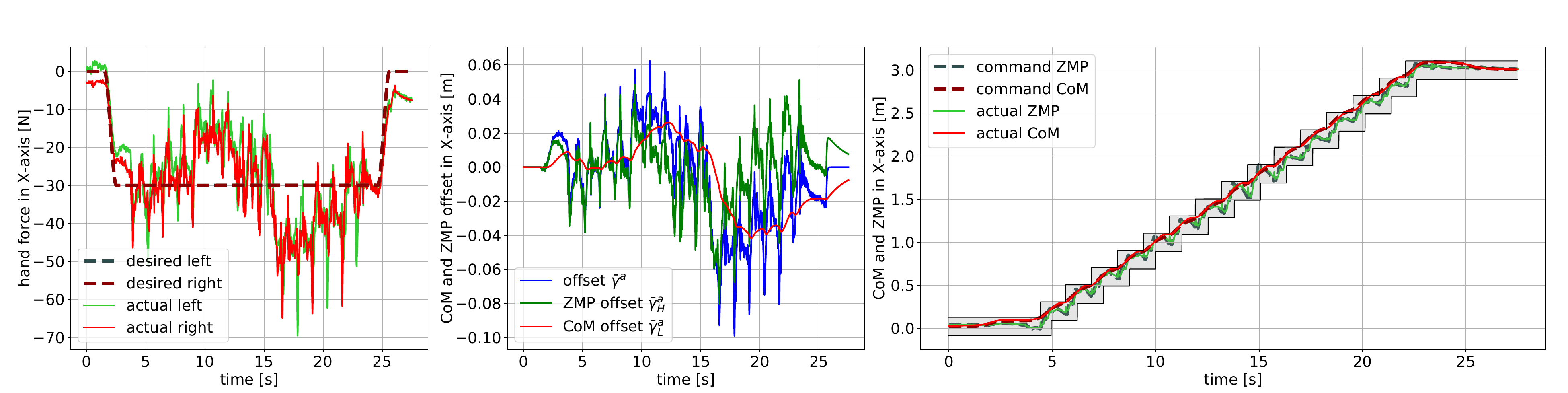}\\
  \begin{minipage}{0.26\linewidth}
    \begin{center} \footnotesize \hspace{6mm} (A) Hand forces \end{center}
  \end{minipage}
  \begin{minipage}{0.26\linewidth}
    \begin{center} \footnotesize \hspace{6mm} (B) Offsets due to external forces \end{center}
  \end{minipage}
  \begin{minipage}{0.46\linewidth}
    \begin{center} \footnotesize \hspace{6mm} (C) CoM and ZMP \end{center}
  \end{minipage}
  \vspace{-1.5mm}
  \caption{Cart pushing operation by HRP-2Kai.}
  \vspace{-4mm}
  \label{fig:exp-cart}
\end{figure*}

\figref{fig:exp-bobbin} shows the rolling operation of a large bobbin (bobbins are found in factories fabricating metal wires, papering, tissues...) by HRP-5P~\cite{HRP5P:Kaneko:RAL2019}.
Such bobbins are generally large and heavy (about 1.3~m in diameter and can reach up to 140~kg in weight), and require a large friction force to curve and turn, making them one of the most difficult objects that humanoid robots have ever handled.
The actual manipulation forces and ZMP were measured from the 6-axis force sensors mounted on the robot's wrists and ankles.
As shown in (A) and (B) in \figref{fig:exp-bobbin}, the robot moved the bobbin by straight-line and turning motions.
Although the setting of the hand forces was not necessary for the straight-line motions on flat grounds due to the small rolling friction, the turning motion could require the lateral hand forces of about 40~N or more for each hand. 
As shown in (C) and (D), the robot turned the bobbin by exerting the hand forces required to move the object while stepping to the side and maintaining its balance.

\begin{figure}[tpb]
  \begin{center}
    \includegraphics[width=0.49\columnwidth]{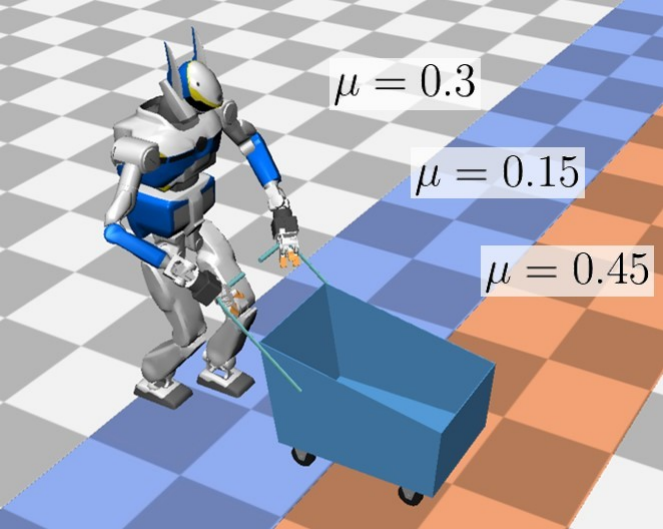}
    \hfill
    \includegraphics[width=0.49\columnwidth]{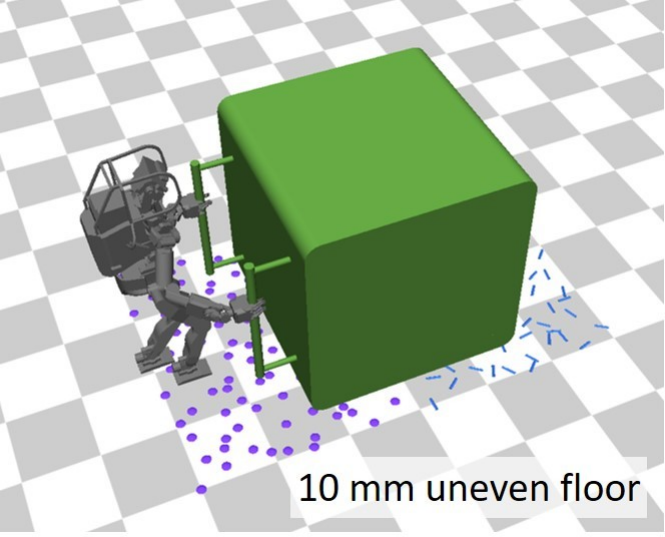}\\
    \begin{minipage}{0.49\linewidth}
      \begin{center} \footnotesize (A) Floor with ununiform friction \end{center}
    \end{minipage}
    \hfill
    \begin{minipage}{0.49\linewidth}
      \begin{center} \footnotesize (B) Uneven floor \end{center}
    \end{minipage}
    \vspace{-6mm}
    \caption{Pushing operation on the floor with model error.}
    \vspace{-4mm}
    \label{fig:exp-sim}
  \end{center}
\end{figure}

\figref{fig:exp-cart} and \figref{fig:exp-sim}~(A) show the cart pushing operation by HRP-2Kai on the floor with changes in friction.
On a floor with a different friction coefficient than expected, there is an error between the desired forces and the actual forces of the hands. However, the CoM strategy and the ZMP strategy in the ST can compensate for the error, and the walking while pushing can be performed robustly.
Additionally, as shown in \figref{fig:exp-sim}~(B), simulation experiments were performed in various scenarios.
See the accompanying video for details.

\section{Conclusion}

In this letter, we propose a control method for humanoid loco-manipulations.
First, we formulate the bipedal dynamics by accounting external manipulation forces, and then introduce a loco-manipulation pattern generator to track the reference trajectory of external forces inherent to manipulation of objects by a humanoid robot.
Second, we formulate a stabilization control that separates the discrepancies between planned and actual manipulation forces in the frequency domain and compensates for them by adjusting CoM and ZMP.
We have shown the effectiveness of our controller by applying it to various loco-manipulation motions such as rolling an object by a humanoid robot.

Future challenges include (i) accurate manipulation of large and heavy objects by integrating full experimental loco-manipulation and assembly operations using object visual tracking and SLAM, and (ii) reaching human-speed performance in such operations.





\section*{Acknowledgment}

We thank Kenji Kaneko and Hiroshi Kaminaga of CNRS-AIST JRL for robot hardware support.

\bibliographystyle{IEEEtran}
\bibliography{main.bib}

\end{document}